# CANDY: Conditional Adversarial Networks based Fully End-to-End System for Single Image Haze Removal

Kunal Swami and Saikat Kumar Das

(Abstract) Single image haze removal is a very challenging and ill-posed problem. The existing haze removal methods in literature, including the recently introduced deep learning methods, model the problem of haze removal as that of estimating intermediate parameters, viz., scene transmission map and atmospheric light. These are used to compute the haze-free image from the hazy input image. Such an approach only focuses on accurate estimation of intermediate parameters, while the aesthetic quality of the haze-free image is unaccounted for in the optimization framework. Thus, errors in the estimation of intermediate parameters often lead to generation of inferior quality haze-free images. In this paper, we present *CANDY* (*C*onditional *A*dversarial *N*etworks based *D*ehazing of haz*Y* images), a fully end-to-end model which directly generates a clean haze-free image from a hazy input image. *CANDY* also incorporates the visual quality of haze-free image into the optimization function; thus, generating a superior quality haze-free image. To the best of our knowledge, this is the first work in literature to propose a fully end-to-end model for single image haze removal. Also, this is the first work to explore the newly introduced concept of generative adversarial networks for the problem of single image haze removal. The proposed model *CANDY* was trained on a synthetically created haze image dataset, while evaluation was performed on challenging synthetic as well as real haze image datasets. The extensive evaluation and comparison results of *CANDY* reveal that it significantly outperforms existing state-of-the-art haze removal methods in literature, both quantitatively as well as qualitatively.

## 1. INTRODUCTION

Images captured under bad weather conditions, such as fog, mist or haze suffer from problems, such as limited visibility, poor contrast, faded colors and loss of sharpness. These artifacts not only significantly deteriorate the aesthetic beauty of captured scenes, but also occlude salient and important regions in images. In presence of fog or haze particles, the original irradiance received by camera from the scene point gets attenuated along the line-of-sight; this combined with the scattering of the atmospheric light produces a hazy image [1]. The degree of effect at each pixel of the image depends on the depth of the corresponding scene point from the camera. The attenuation phenomenon causes the original scene point irradiance to decrease with increasing depth, while the atmospheric light component increases with increasing depth. Therefore, the traditional space invariant image processing techniques, such as contrast enhancement and image histogram equalization fail to correctly remove spatially variant haze degradation from images.

The mathematical model for the formation of a hazy image which was first formulated by Koschmieder [2] and is used by almost all the methods in literature is as follows:

$$I(x) = J(x)t(x) + \alpha(1 - t(x)) \quad (1)$$

Here, $I(x)$ is the observed hazy image, $J(x)$ is the actual scene irradiance, $t(x)$ is the scene transmission map, $\alpha$ is the atmospheric light and $x$ denotes an individual pixel location in the image. The scene transmission, as stated earlier, is a function of depth and is given by:

$$t(x) = e^{-\beta d(x)} \quad (2)$$

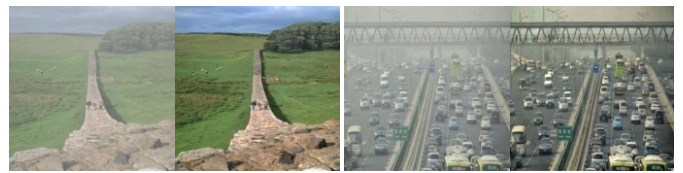

(a) Synthetic haze example     (b) Real haze example

**Fig. 1. Example results of *CANDY* on synthetic and real images.**

Here, $d(x)$ is the depth of the scene point corresponding to the pixel location $x$ and $\beta$ is the scattering coefficient of the atmosphere which represents the ability of a unit volume of atmosphere to scatter light in all directions [3]. The scene transmission decreases exponentially with increasing distance of the scene point from the camera. It is easy to observe that Eq. (1) contains three unknowns, viz., $J(x)$, $t(x)$ and $\alpha$, which makes the problem of finding $J(x)$ under constrained. As a result, single image haze removal is a very challenging and ill-posed problem.

The problem of haze removal is important because it improves the overall aesthetic beauty of the image by restoring the original scene contrast and color vividness. Smartphone cameras are being used ubiquitously to capture images of wide variety of scenes. Bad weather conditions, such as fog or haze, deteriorate the overall aesthetic quality of captured images and often render them useless. Existing fog or haze removal techniques in smartphones are simply based on contrast enhancement or image histogram equalization, which fail to generate visually pleasing results in challenging real scenarios. A robust haze removal technique is a very important feature for Samsung smartphone cameras from commercial point of view. Apart from consumer devices, haze removal is an important pre-processing step in most image processing and computer vision systems which work on the basis of the assumption that input images are free of artifacts due to unfavorable weather conditions. Haze removal is also important in outdoor video surveillance cameras.

In literature, the existing haze removal methods can be broadly classified into following three categories:

a. **Multiple Images:** Methods which use multiple images captured under different weather or polarization conditions [1], [3-6].
b. **Single Image:** Methods which perform haze removal using a single degraded hazy image [8-26].
c. **Deep Learning:** Recently proposed single image haze removal methods in [30- 32] use deep learning.

*A common and key step in all the existing haze removal methods is to estimate intermediate scene transmission map ($t(x)$) and atmospheric light component ($\alpha$).* The estimated scene transmission and atmospheric light are then substituted in Eq. (1) to obtain the original haze-free image $J(x)$. The scene transmission and atmospheric light component are estimated independently, joint optimal estimation of these intermediate parameters is not performed by any of the existing works. The atmospheric light component is either calculated from the estimated transmission map using ad-hoc empirical rules or it is regarded as a global constant which is not true. Moreover, some methods [30] require one or a few parameters to be set manually. Following are the inherent drawbacks of the approach adopted by existing methods in literature:

a. The approach adopted by existing methods does not consider visual quality of the generated haze-free image into the optimization framework. In other words, existing methods don't directly focus on generating high-quality haze-free image; rather the focus is on accurately estimating the intermediate parameters.
b. As existing methods focus only on estimation of scene transmission and atmospheric light, inaccuracies in the estimation of these intermediate parameters often lead to erroneous or inferior quality haze removal.

In this paper, we address the aforementioned limitations of existing haze removal methods. We present *CANDY—Conditional Adversarial Networks based Dehazing of hazY* images, a novel fully end-to-end model which directly generates haze-free image from a hazy input image. The proposed model also incorporates visual quality of the generated haze-free image into the optimization function. In contrast to existing deep learning based methods [30-32] which learn an end-to-end mapping from a hazy image to its transmission map, *CANDY* is a fully end-to-end system which learns the complete atmospheric model and directly generates superior quality haze-free image from a hazy input image (see Fig. 1(a) and 1(b)). *To the best of our knowledge, this is the first work in literature to propose a fully end-to-end model for single image haze removal. At the same time, this is also the first work to explore the newly introduced concept of generative adversarial networks* [33] *for single image haze removal problem.*

The outline of this paper is as follows: a review of existing haze removal techniques in literature and the major contributions of this work are presented in Section 2. Section 3 explains the architecture of our proposed deep learning model *CANDY*. Section 4 and 5 are devoted to training and experimentation details, while Section 6 presents our results and their comparison with existing state-of-the-art methods. Finally, concluding remarks are given in Section 7.

## 2. RELATED WORK

**Multiple Images:** Early methods by Narasimhan and Nayar [1], [3] and [4] performed haze removal by utilizing multiple images captured under different weather conditions, whereas methods presented in [5] and [6] used multiple images with different degrees of polarization in order to produce a haze-free image. Another method proposed by Schaul *et al*. [7] used an additional near-infrared image of the scene to perform haze removal. The problem with these methods is their limited practical applicability because capturing multiple images of a scene is difficult and not always possible, especially when the scene is dynamic. Moreover, the assumption of static scene is too strong to hold true in many real scenarios.

**Single Image:** The early single image haze removal methods proposed by Narasimhan and Nayar [8] and Kopf *et al*. [9] relied on user supplied information about the scene structure. Hautière *et al*. [10] proposed a method to remove haze from images captured from a moving vehicle camera. Fattal [11] used Independent Component Analysis based method to estimate albedo and transmission of a scene based on the assumption that they are locally uncorrelated. However, extensive computation time and unsatisfactory performance in dense haze limits the applicability of their method. Tan [12] proposed a method which maximizes local contrast for haze removal; it was based on the observation that haze-free images have higher contrast compared to its hazy version. He *et al*. [13] proposed a novel dark channel prior based haze removal method. Dark channel is composed of lowest intensity of each pixel across three image channels and in absence of haze the channel remains completely dark or has very little intensity. However, this method fails to work in scenes where major portion is covered by atmospheric light or similar object, such as sky. Tarel and Hautière [14] addressed the problem of large computational time in [11-13] using the median of median filter. In [15] and [16], authors proposed a method using Factorial Markov Random Field to jointly estimate scene albedo and depth assuming that they are statistically independent latent layers. Meng *et al*. [17] addressed the problem of abrupt depth jumps in local image patches by imposing a boundary constraint on the transmission function; earlier methods assumed that all pixels in a local image patch share similar depth value. Ancuti and Ancuti [18] were first to propose a fusion based strategy that derives two hazy images from the hazy input image by applying white-balance and contrast enhancement. These images were blended effectively in a per-pixel fashion by computing weight maps based on luminance, chrominance and salient region features. Sulami et al. [19] focused on accurately estimating the atmospheric light vector by exploiting the abundance of small image patches in which scene transmission and albedo are constant. Mai et al. [20] proposed a simple neural network to predict transmission value at each pixel of the input image. Fattal [21] proposed a haze removal algorithm which was based on the observation of a generic regularity in natural images in which pixels of small image patches exhibit one dimensional distribution in *RGB* color space. An investigation of several haze-relevant features was performed by Tang et al. [22]. They trained a Random Forest classifier to predict transmission values for an input image patch; it was found that dark channel prior is the most important clue to estimate the scene transmission map. In another method by Zhu et al. [23], authors proposed a new color attenuation prior based method to estimate the transmission map from a hazy input image. Li and Zheng [24] used a weighted guided image filter to decompose the dark channel of the hazy input image into base and detail layer and used the base layer of dark channel to estimate the transmission map. This way, they could preserve prominent edges after haze removal. The methods presented in [25], [26] focused on enhancing the works discussed so far. Ma et al. [27] presented an evaluation study of single image haze removal methods discussed so far. Another such study was also presented by Ancuti et al. [28] using a challenging dataset derived from *NYU* depth dataset [29]. In both these works, it was observed that none of the existing haze removal methods could produce high quality haze-free images without artifacts.

**Deep Learning:** Recently the problem of haze removal was also addressed using deep learning approaches [30-32]. These methods were also based on the approach of estimating the intermediate transmission map and atmospheric light. Ren et al. [30] presented a multi-scale convolutional neural network based approach where they first predicted a coarse holistic transmission map which was later refined using a different fine-scale network. This method requires manual tuning of parameter for gamma correction based on the density of haze in the input image. Cai et al. [32] presented another deep convolutional network model called *DehazeNet* which learned a direct mapping from a hazy input image to the scene transmission map.

It can be inferred from the above discussion that the key idea of existing haze removal methods is to estimate the scene transmission map and atmospheric light from a hazy input image. The haze-free image is then computed in a separate step. None of the existing methods perform joint optimal estimation of these intermediate parameters nor do they incorporate aesthetic quality of the generated haze-free image into the optimization framework. In this work, we address the aforementioned drawbacks and present a novel fully end-to-end deep learning model to directly generate a high quality haze-free image from a hazy input image.

In summary, following are the major **contributions** of this work:

- This is one of the first works in literature to propose a fully end-to-end model for single image haze removal problem [1]. The proposed model *CANDY* directly generates a clean haze-free image from a hazy input image.
- This is also the first work to explore generative adversarial networks for the single image haze removal problem. The discriminator network ensures that generated haze-free image looks indistinguishable from original haze-free image. It is shown in experimental results that incorporating adversarial loss into the optimization function considerably improves the quality of generated haze-free images.
- As generative adversarial networks are difficult to train and the generated images are susceptible to considerable artifacts, we conduct experiments with combination of different types of losses, including the recently introduced feature reconstruction loss [34] for improving the quality of generated haze-free images.
- An extensive evaluation and comparison of the proposed model *CANDY* on challenging synthetic as well as real haze image datasets reveals that it significantly outperforms the existing state-of-the-art methods in literature [13], [14], [17], [19], [30] and [32]; both quantitatively as well as qualitatively.

## 3. MODEL ARCHITECTURE

In this work, we formulate the problem of single image haze removal as a problem of generating a high quality haze-free image from a degraded hazy input image. As stated earlier, it is clear from Eq. (1), that the problem of single image haze removal is under constrained. In order to tackle this under constrained and non-linear problem of mapping, we propose a novel deep conditional generative adversarial network called *CANDY*. The model architecture comprises of two deep convolutional neural network modules, viz., a generator ($G$) and a discriminator ($D$), who's combined efforts lead to generation of high-quality haze-free image from a degraded hazy input image. Generative Adversarial Networks (*GANs*) were first introduced by Goodfellow et al. [33]; Mirza and Osindero [35] later presented *GAN*s in conditional setting called conditional generative adversarial networks (*CGAN*s). After this, *CGAN*s and their variants have been successfully applied in various image generation and transformation problems, such as [36-38].

In brief, *GAN*s are generative models which learn a mapping from random noise vector $z$ to output image $y$. Mathematically, $G : z \rightarrow y$. In contrast, *CGAN*s learn a mapping from input image $x$ and random noise vector $z$ to output image $y$. Mathematically, $G : \{x, z\} \rightarrow y$. The generator network $G$ is trained to generate output images which are indistinguishable from real images to an adversary, called discriminator network $D$. The discriminator network $D$ is trained to correctly distinguish between *real* images and *fake* images synthesized by generator network $G$. This idea is similar to a two player minimax game [33]. In best scenario, both *G* and *D* are expected to reach a saddle point at which none can improve further. The objective function of *CGAN*s can be mathematically expressed as:

$$L_{CGAN}(G,D) = \mathbb{E}_{x,y\sim p_{data}(x,y)}[\log D(x,y)] + \mathbb{E}_{x\sim p_{data}(x),z\sim p_z(z)}[\log(1 - D(x,G(x,z)))] \quad (3)$$

---

[1] This work was performed in April 2017, when such an end-to-end approach was not *public*. We acknowledge that Boyi Li et al. [48] were first to explore a complete end-to-end approach, whose work "AOD-Net" was accepted in ICCV 2017 (submission in March 2018)

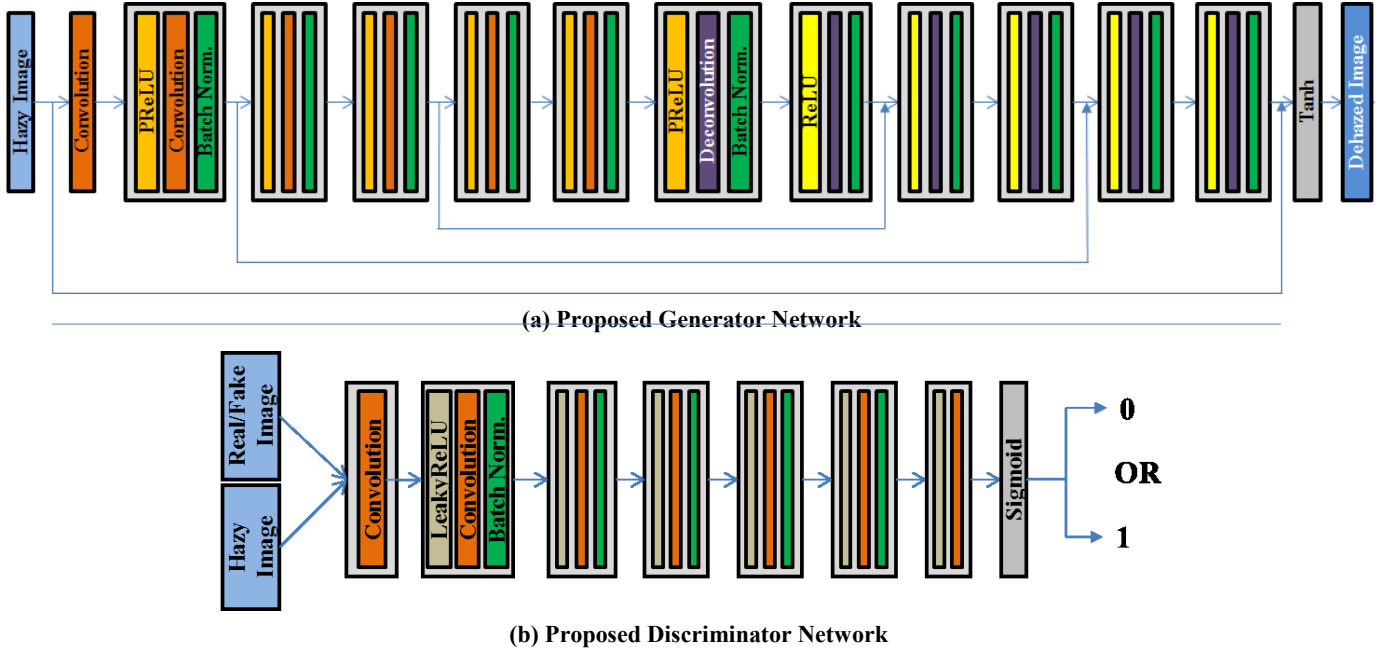

**Fig. 2.** Architecture of the proposed model *CANDY* (best view in color).

Here, the generator *G* tries to minimize this objective function against the discriminator *D*, which in turn tries to maximize it. Therefore, the training objective of *CGAN*s is expressed as:

$$\arg min_G \, max_D \, L_{CGAN}(G,D) \quad (4)$$

The generator and discriminator networks of the proposed model *CANDY* have been designed on the basis of reported experimental findings and recommendations by Radford *et al.* [36] and Isola *et al.* [37]. The model architectures are explained in detail in the following subsections.

### 3.1. Generator Network

The proposed generator network as shown in Fig. 2(a) is a fully convolutional deep neural network. It comprises of six convolution layers followed by six deconvolution layers and a *Tanh* (hyperbolic tangent function) output layer at the end. Each convolution layer is followed by batch-normalization and *PReLU* (parametric rectified linear unit) activation layer, except the first convolution layer where batch-normalization is not applied. Each deconvolution layer is followed by *ReLU* (rectified linear unit) activation layer and batch normalization. Unlike *ReLU*, *PReLU* does not ignore negative gradient values, hence, preventing gradient saturation. Mathematically, Eq. (5) describes the *ReLU* operation and Eq. (6) describes the *PReLU* operation. The parameter λ (coefficient of leakage) controls the magnitude of negative gradient and is learned along with other network parameters.

$$r(x) = \max(0,x) \quad (5)$$

$$pr(x) = \max(0,x) - \lambda * \max(0,-x) \quad (6)$$

It can be inferred from Fig. 2(a) that the generator architecture is symmetric in nature with convolution layers acting as *encoder* (feature extraction), while deconvolution layers acting as *decoder* (image recovery). Rather than using convolution layers throughout the generator network, we preferred to use convolution and deconvolution layers because deconvolution layers have been proven to be better in recovering finer image details [36]. Each of convolution and deconvolution layers use filter size of 3 x 3 and generate 64 feature maps. The size of feature maps is kept same as input image size by setting stride of length 1 and zero-padding of length 1. Down-sampling is not performed as it leads to loss of important features in image [37]. In addition to the existing design features in our generator network, skip connections [39] are added to tackle the problem of vanishing gradients. Skip connections are added after every two convolution layers to their corresponding deconvolution layers, the feature maps from convolution layer are element-wise summed to deconvolution layer feature maps. Another benefit of skip connections is that they pass image details from convolution layers directly to the deconvolution layers; thus, promoting recovery of original image details. The input to the proposed generator network *G* is a hazy image and the output is a high quality haze-free image. The generator is fully convolutional and accepts input image of any size.

## 3.2. Discriminator Network

The proposed discriminator network as shown in Fig. 2(b) is also a fully convolutional deep neural network. It contains a stack of seven convolution layers, where each convolution layer (except first) is followed by batch-normalization and Leaky *ReLU* activation. In case of Leaky *ReLU* operation, the parameter λ in Eq. (6) is fixed and set manually, in our case we set λ = 0.2. The input to the discriminator network is a pair of images concatenated along the channel axis. The input image pairs are of two types:

a. **Real pair (positive example):** Hazy image and ground-truth haze-free image
b. **Synthesized pair (negative example):** Hazy image and haze-free image synthesized by *G*

The last layer in the discriminator network is a sigmoid layer which outputs the probability of the input image pair to be real (1) or fake (0). The discriminator network is alternatively fed with positive and negative image pairs. This way, the discriminator *D* learns to distinguish between real haze-free images (ground-truth) and synthesized haze-free images generated by the generator network *G*. The output of the discriminator network *D* constitutes the *adversarial loss*, which is used to train the generator network *G* whose goal is to fool the discriminator by generating haze-free images which are indistinguishable from their original ground-truth haze-free images. The convolution kernel size used in the discriminator network is 3 x 3 with stride of length 2 and zero-padding of length 1. The number of feature maps doubles after every two convolution layers, starting from 48 after the first convolution layer. The last convolution layer outputs a single feature map which is input to sigmoid layer.

## 4. TRAINING OBJECTIVE

In Section 3, a brief explanation about *CGAN*s and its training objective was presented. In the proposed model *CANDY*, however, we don't provide noise vector *z* along with the input image to the generator, as conditional generative adversarial networks have been found to ignore random noise *z* [37]. Hence, the modified objective function for the proposed model *CANDY* can be stated as follows:

$$L_{CANDY}(G,D) = \mathbb{E}_{x,y \sim p_{data}(x,y)}[\log D(x,y)] \\ + \mathbb{E}_{x \sim p_{data}(x)}[\log(1 - D(x, G(x)))] \quad (7)$$

As already stated in Section 3, here *x* is the hazy input image and *y* is the ground-truth haze-free image. *GAN*s are highly unstable to train and the generated images often contain artifacts. Hence, apart from the adversarial loss (discriminator output), we propose and experiment with combinations of different types of losses which are explained in following subsections.

### 4.1. Content Loss

To the original training objective in Eq. (4), we also add content loss (pixel-based loss). Traditional content loss functions include $L_1$ and $L_2$ distances. The benefit of using content loss is that it encourages the generator to generate images which are closer to ground truth images in $L_1$ (or $L_2$) sense; in other words, it helps to minimize pixel-level differences between the generated image and the ground-truth image. The Euclidean ($L_2$) loss between ground-truth haze-free image *y* and the generated haze-free image *G(x)* (where *x* is the hazy input image) is given by:

$$L_2 = \|y - G(x)\|_2 \quad (8)$$

However, it is well known that $L_2$ loss enforces strong penalty due to the squared terms and hence, susceptible to blurring [37]. Therefore, we also experiment with Smooth $L_1$ loss ($L_{s1}$) which is less penalizing and is defined as follows:

$$L_{s1}(y, G(x)) = \begin{cases} 0.5d^2, & \|d\|_1 < 1 \\ \|d\|_1 - 0.5, & otherwise \end{cases} \quad (9)$$

Here,

$$d = y - G(x) \quad (10)$$

### 4.2. Feature Reconstruction Loss

Johnson *et al*. [34] observed that rather than only encouraging the pixels of the output image to match with the ground-truth image using content loss, it is beneficial to also minimize the differences between their high-level feature representations by a convolutional network. They called this loss as *feature reconstruction loss* and found that it significantly improves the output image quality for super-resolution and image stylization tasks. The basic intuition behind this concept is that minimizing the difference between high-level feature representations (which encode high-level image semantics) helps to preserve the overall spatial structure and semantic content of generated image.

Let $\varphi_i(x)$ denote the activation (feature map) of $i^{th}$ layer of a convolutional neural network $\varphi$, therefore, our feature construction loss is defined as follows:

$$L_{feature}(y, G(x)) = \|\varphi_i(y) - \varphi_i(G(x))\|_2 \quad (11)$$

In our case, we use a pre-trained *VGG-Net* model [34] for calculating the feature reconstruction loss between the ground-truth and generated haze-free image. We experiment with different layers of *VGG-Net* for feature reconstruction loss by setting value of *i* to one of 9 (*relu*2_2), 16 (*relu*3_3) and 23 (*relu*4_3).

### 4.3. Final Loss Function

The final loss function *L* of the proposed model which is minimized during training by the optimization framework is composed of losses described in Eq. (4), (8), (9) and (11).

$$L = \arg\min_G \max_D L_{CANDY}(G,D) \\ + L_2(y, G(x)) + L_{s1}(y, G(x)) \\ + L_{feature}(y, G(x)) \quad (12)$$

The generator network learns to generate haze-free images by minimizing the loss function described in Eq. (12). The weights of different types of losses were determined empirically using validation dataset.

## 5. EXPERIMENTATION DETAILS

### 5.1. Dataset Creation

As there is no haze image dataset in literature, similar to existing methods we synthetically created our own haze image dataset. For this purpose, we used popular *Make3D* depth dataset [40], [41] and *BSDS500* dataset [42]. Some of the existing methods [30], [32] have used *NYU* depth dataset [29] but this dataset only contains indoor images, which is not the most likely place to find weather conditions like fog or haze (except smoke). In order to create hazy images, a synthetically generated scene transmission map is required. To synthesize haze which looks realistic, we generated depth map for all the images using the state-of-the-art depth map estimation method of Liu *et al*. [43]. The depth map of an image was then used to synthesize hazy images as follows (please refer to Eq. (1) and (2)): Given an input image and its depth map, we randomly sample atmospheric light $\alpha$ and extinction coefficient $\beta$, such that $\alpha = [k\ k\ k]$ (one for each channel), where $k \in [0.7, 1]$ and $\beta \in [0.5, 1.5]$. These values are substituted in Eq. (1) and (2) to obtain a synthetic hazy image.

### 5.2. Training and Evaluation Datasets

First, random 500 and 200 images were selected from *Make3D* and *BSDS500* datasets respectively. For each of these 700 images, 3 synthetic hazy images (each with random $\beta$ and $\alpha$) were generated using the procedure in Section 5.1. Thus, the training dataset derived from *Make3D* and *BSDS500* datasets contained 2100 hazy and ground-truth haze-free image pairs. Apart from this, a small validation dataset of 40 hazy and ground-truth haze-free image pairs was used for tracking model performance and empirically determining the hyper-parameters of the proposed model.

Testing was performed on synthetic as well as real haze image datasets. Two different synthetic test datasets, viz., *Test-Synthetic-A* and *Test-Synthetic-B* were created. Test-Synthetic-A contained 90 hazy and ground-truth haze-free image pairs from *Make3D* and *BSDS500* datasets, whereas Test-Synthetic-B contained 23 hazy and ground-truth haze-free image pairs from a completely different *Middlebury* dataset [44]. The real haze image test dataset, *Test-Real-500* contained challenging 500 real hazy images.

### 5.3. Training Details

The experiment was performed using Torch framework [45] with NVIDIA GTX Titan X GPU. The model was trained with Adam optimization method [46] using learning rate $2 \times 10^{-4}$ and momentum term 0.5. The model was trained till 1000 epochs using batch-size of 4. The size of the images used to train and evaluate our model was 256 x 256 x 3. However, as stated earlier, the proposed generator network can accept input image of any size. The model was evaluated on validation dataset after every 50 training epochs.

### 5.4. Evaluation Criteria and Metrics

The proposed model was evaluated quantitatively using the two synthetic test datasets described in Section 5.2 which contain pairs of synthetic hazy and ground-truth haze-free images. The Peak Signal to Noise Ratio (*PSNR*) and Structured Similarity Index Measurement (*SSIM*) [47] metrics were used to quantitatively measure similarity of generated haze-free images with the corresponding ground-truth haze-free images. *SSIM* metric is used because it is considered more close to how humans perceive visual similarities and differences in images [47].

Apart from quantitative evaluation, qualitative evaluation is also performed using both the synthetic as well as real test datasets described in Section 5.2.

### 5.5. Comparisons with Baseline Model and Variants

In order to prove the effectiveness of adversarial training, we first trained a baseline model *GEN*. The architecture of the model *GEN* is exactly same as the proposed generator model (see Fig. 2(a)); however, it was trained only using the $L_2$ and feature reconstruction loss function, adversarial cost was *not* included. GEN was trained till 1000 epochs. The version of model *GEN* which was trained till 500 epochs (*GEN*-500) was used to initialize weights of *CANDY*. This was done to prevent *CANDY* from converging into a local minimum because *GAN*s are highly unstable and difficult to train. Also, we experimented and evaluated different versions of *CANDY*. The different configurations of models experimented and evaluated in this work are listed below:

a. **GEN**: Only generator network was trained with $L_2$ loss and feature reconstruction loss with $i$=9. This model was trained from scratch till 1000 epochs.
b. **CANDY-L1-9P**: Initialized with weights of *GEN*-500, trained using smooth $L_1$ loss and feature reconstruction loss with $i$=9. This model was trained till 500 epochs.
c. **CANDY-L2-9P**: Initialized with weights of *GEN*-500, trained using $L_2$ loss and feature reconstruction loss with $i$=9. This model was trained till 500 epochs.
d. **CANDY-L1-23P**: Initialized with weights of *GEN*-500, trained using smooth $L_1$ loss and feature reconstruction loss with $i$=23. This model was trained till 500 epochs.
e. **CANDY-L2-23P**: Initialized with weights of *GEN*-500, trained using $L_2$ loss and feature reconstruction loss with $i$=23. This model was trained till 500 epochs.

### 5.6. Comparisons with Existing State-of-the-Art

The results of the proposed model *CANDY* were extensively evaluated and compared with existing state-of-the-art methods in literature. The methods which were compared with *CANDY* are: He *et al*. [13], Tarel *et al*. [14], Meng *et al*. [17], Sulami *et al*. [19], Ren *et al*. [30] and Cai *et al*. (*DehazeNet*) [32]. The original code provide by the respective authors was used to perform evaluation and comparison. Only He *et al*.'s [13] code was implemented.

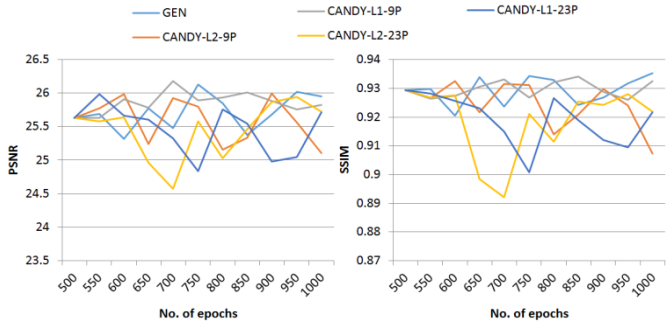

**Fig. 3.** Quantitative comparison of different model configurations on validation dataset (best view in color).

## 6. RESULTS AND DISCUSSION

We evaluated all the different model configurations described in Section 5.5 on the validation dataset and the quantitative results obtained are shown graphically in Fig. 3. The evaluation and comparison was performed in terms of *SSIM* and *PSNR* metrics. *GEN* and *CANDY-L1-9P* were top performers on the validation dataset with *SSIM* score of 0.9353, 0.9325 respectively and *PSNR* score of 25.9536, 25.8277 respectively. The combined average *SSIM* and *PSNR* scores of other model configurations were 0.9169 and 25.5160 respectively. It can also be inferred from Fig. 3 that *GEN* and *CANDY-L1-9P* consistently performed better than other model configurations. A subjective evaluation of results was also performed and it was in agreement with the quantitative results. $L_2$ loss was found to be less effective for adversarial training; it showed high variations and often led to considerable artifacts in the generated images. Smooth $L_1$ loss was found to be more effective and it decreased steadily during the adversarial training. The same observation can also be made from graphs in Fig. 3, where variants of *CANDY* trained with $L_1$ loss performed better than other variants trained with $L_2$ loss. Also, feature reconstruction loss using lower layers of *VGG-Net* (in our case, $i$=9) lead to better quality results. This can be attributed to the fact that while higher layers of a convolutional network preserve the overall spatial structure of an image, they do not preserve the original image texture and color [46]. Fig. 4 demonstrates the problems observed on one of the sample image from validation dataset when higher layers of *VGG-Net* were used for calculating the feature reconstruction loss. Based on the results obtained on validation dataset, we select the baseline model *GEN* and the adversarial model *CANDY-L1-9P* (hereinafter, referred to as *CANDY*) as our final models.

The conditional adversarial model *CANDY* and the baseline model *GEN* are finally evaluated on the two synthetic test datasets: Test-Synthetic-A and Test-Synthetic-B. Table 1 shows the quantitative results of *CANDY* against the baseline model *GEN*. The results obtained with *CANDY* are considerably better than *GEN*, thus, proving the effectiveness of adversarial training.

The results of the proposed model *CANDY* are finally compared against state-of-the-art methods in literature using both synthetic as well as real haze image datasets which are described in Section 5.2. Table 2 shows the comparison of quantitative results obtained with *CANDY* and existing state-of-the-art methods on Test-Synthetic-A and Test-Synthetic-B datasets respectively. It can be observed that results generated by *CANDY* are significantly better, both in terms of *SSIM* and *PSNR* metric. The performance of *CANDY* on both the synthetic test datasets is similar and significantly better than the state-of-the-art; thus, proving its superiority.

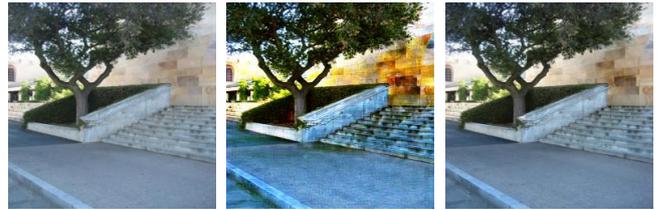

(a) Ground-truth  (b) *L2-23P* result  (b) *L1-9P* result

**Fig. 4.** Effect of different losses on a sample validation image

**Table 1.** *GEN* vs. *CANDY* on synthetic test datasets

|  | Test-Synthetic-A | | Test-Synthetic-B | |
|---|---|---|---|---|
|  | *GEN* | *CANDY* | *GEN* | *CANDY* |
| SSIM | **0.9286** | 0.9285 | 0.9232 | **0.9313** |
| PSNR | 25.4273 | **25.6454** | 23.7934 | **24.11** |

Fig. 5 shows qualitative results of *CANDY* on synthetic hazy images and compares it with the results of existing state-of-the-art methods. Fig. 6 demonstrates the qualitative results of *CANDY* and other methods on some of the very challenging real hazy images from Test-Real-500 dataset. It can be observed from both Fig. 5 and 6 that *DehazeNet* [32] and method of Ren *et al*. [30] fail to completely remove haze from images. It was observed that *DehazeNet* [32] often makes some parts of image darker, whereas Ren *et al*.'s method often significantly increases color saturation in images. The results of other methods [13], [17], [19], especially Tarel *et al*. [14] are not visually pleasing. In contrast, the results generated by *CANDY* from synthetic hazy images in Fig. 5 are very close to the actual ground-truth haze-free images. In Fig. 6, we have shown dehazing results of some of the very challenging real hazy images to demonstrate superiority of *CANDY* over existing methods. In Fig. 6(a), *CANDY* is able to remove maximum haze, while also enhancing colors and sharpness of the image, for e.g., the tree and grass in the image look sharper with enhanced colors. *DehazeNet* [32] and other methods fail to completely remove haze; their results look blurry and contain color artifacts. In Fig. 6(b) as well, *CANDY* was able to remove maximum haze and the generated image looks sharper with improved color saturation. The results of other methods still contain haze, in addition to color artifacts and blur.

In this work, we have additionally evaluated the proposed model *CANDY* on real hazy images captured during night time. Fig. 7 shows the qualitative results of *CANDY* on night time hazy images in comparison with existing state-of-the-art. Although *CANDY* was trained on daytime hazy images, we are pleased to find that it generates good quality results even in case of hazy images captured at night. It must be noted that existing state-of-the-art haze removal methods are based on Koschmieder's model in Eq. (1), which holds true only for daytime haze, as it doesn't account for scattering of light from various different light sources during night time.

**Table 2. Quantitative comparison of *CANDY* with existing state-of-the-art methods on synthetic haze image datasets**

| | He [13] CVPR 2009 | Tarel [14] ICCV 2009 | Meng [17] ICCV 2013 | Sulami [19] ICCP 2014 | Ren [30] ECCV 2016 | *DehazeNet* [32] TPAMI 2016 | *CANDY* |
|---|---|---|---|---|---|---|---|
| **Test-Synthetic-A Dataset** | | | | | | | |
| *SSIM* | 0.7861 | 0.7876 | 0.7531 | 0.6918 | 0.8030 | 0.8480 | **0.9285** |
| *PSNR* | 16.0292 | 13.7377 | 17.3138 | 13.6855 | 20.1530 | 21.7724 | **25.6454** |
| **Test-Synthetic-B Dataset** | | | | | | | |
| *SSIM* | 0.8455 | 0.8028 | 0.7957 | 0.7366 | 0.8697 | 0.9056 | **0.9313** |
| *PSNR* | 16.5938 | 13.8395 | 16.3651 | 14.1600 | 19.6637 | 23.6217 | **24.11** |

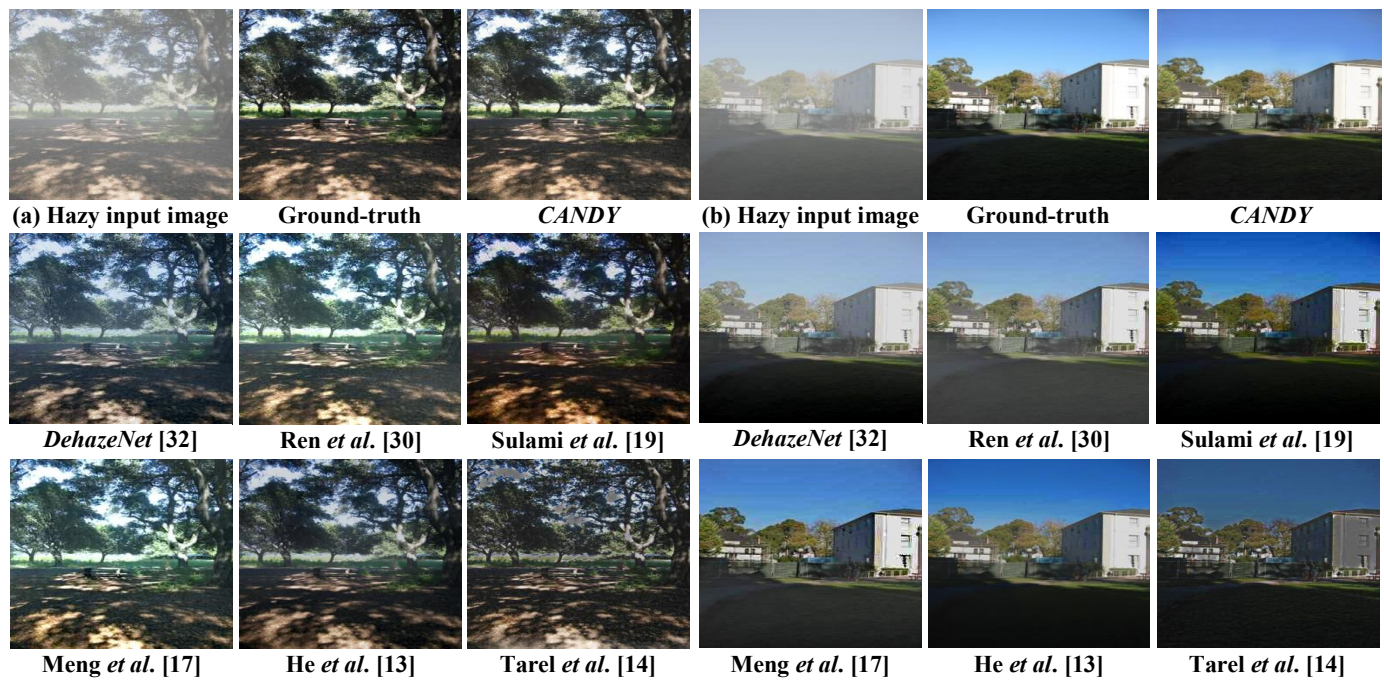

**Fig. 5.** Qualitative comparison of *CANDY* with existing state-of-the-art methods on synthetic hazy images (best view in color).

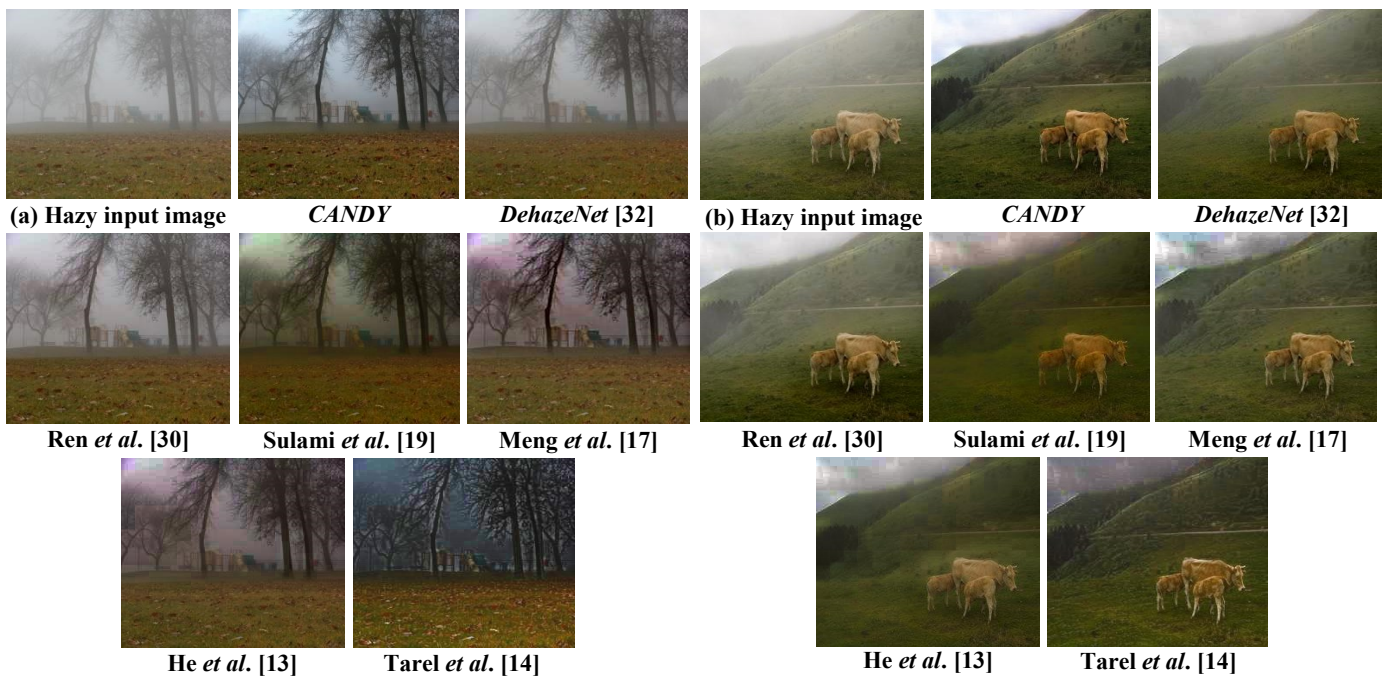

**Fig. 6.** Qualitative comparison of *CANDY* with existing state-of-the-art methods on challenging real hazy images (best view in color).

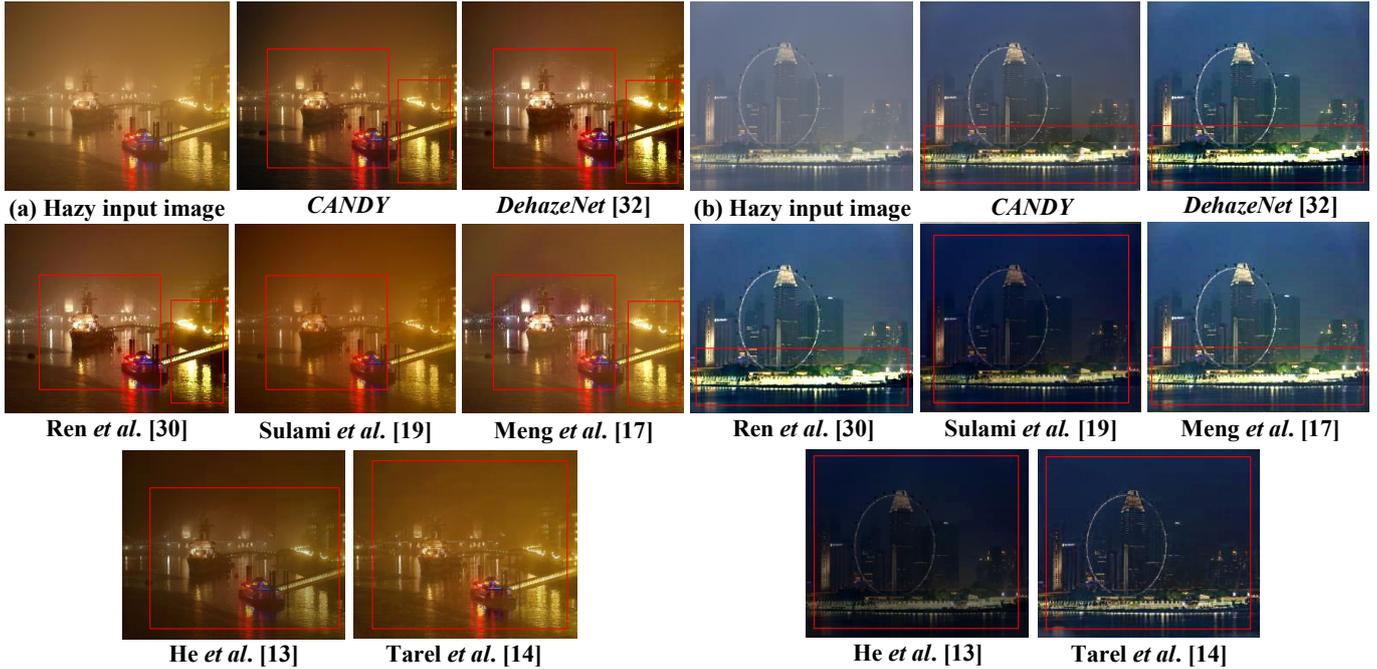

**Fig. 7.** Qualitative comparison of *CANDY* with existing state-of-the-art methods on challenging real night time hazy images. Please note that *CANDY* was trained only using daytime images. Pay attention to regions bounded by red boxes (best view in color).

Consequently, all the existing state-of-the-art methods fail to correctly dehaze night time hazy images. In Fig. 7, it can be observed that *DehazeNet* [37] and other methods fail to completely remove haze and also introduce severe color distortions around the light sources in images. The existing methods also severely boost the glow around the light sources. In contrast, *CANDY* is able to remove significant haze and blurriness from the night time hazy images. It can also be observed from Fig. 7 that *CANDY* handles glow around light sources correctly, while also enhancing the sharpness and color saturation of images. The good performance of *CANDY* in case of night time dehazing can be attributed to the fact that *CANDY* was trained completely end-to-end to learn the complete atmospheric model and generate a superior quality haze-free image.

The quantitative as well as qualitative results clearly demonstrate superiority of *CANDY* over existing state-of-the-art haze removal methods in literature. It is evident from the qualitative results that *CANDY* is not only able to correctly remove haze, but it also restores the sharpness and color vividness of the scene; thus, the haze-free images generated by *CANDY* look visually pleasing compared to the results of existing haze removal methods. This can be attributed to proven image generation capabilities of generative adversarial networks and the design of the optimization function of *CANDY* which accounts for the aesthetic quality of the generated haze-free image. It is evident from the results that *CANDY* is a new state-of-the-art single image haze removal solution.

Lastly, from computational performance point of view, the proposed model *CANDY* takes ~35 ms on a single Nvidia Titan X GPU to generate a haze-free image from a hazy input image of size 256 x 256, whereas it takes ~53 ms for 1024 x 1024 size image. The size of final model is just 3 MB.

## 7. CONCLUSION

In this work, a novel fully end-to-end deep learning model was proposed for addressing the single image haze removal problem. The approach adopted in this work was new and completely different from existing haze removal methods which focus on estimating the scene transmission map and atmospheric light from the hazy input image. The concept of generative adversarial networks was also applied for the first time in literature for haze removal problem. Systematic experiments were performed and the advantage of adversarial training over a baseline convolutional neural network model was demonstrated quantitatively. The proposed model was trained to incorporate the aesthetic quality of the generated haze free image. The quantitative as well as qualitative results demonstrated that the proposed model generates superior quality haze-free images with sharp texture details and improved color saturation; thus, significantly outperforming the existing state-of-the-art methods in literature. As the proposed model was designed and trained to learn the complete atmospheric model, it performed exceptionally well even in case of night time hazy scenes. The model proposed in this work has set a new state-of-the-art for singe image haze removal.